\documentclass{article}
\PassOptionsToPackage{numbers}{natbib}

 \usepackage[preprint]{neurips_2026}

\usepackage[utf8]{inputenc} 
\usepackage[T1]{fontenc}    
\usepackage{hyperref}       
\usepackage{url}            
\usepackage{booktabs}       
\usepackage{amsfonts}       
\usepackage{nicefrac}       
\usepackage{microtype}      
\usepackage{xcolor}         
\usepackage[dvipsnames]{xcolor}
\definecolor{darkgreen}{rgb}{0.0, 0.69, 0.31}
\usepackage{colortbl}
\usepackage{amsmath}    
\usepackage{amssymb}    
\usepackage{mathtools}  
\usepackage{multirow}
\definecolor{lightgray}{gray}{0.9}
\usepackage{graphicx}  
\usepackage{wrapfig}
\usepackage{hyperref}
\graphicspath{{figures/}}
\title{TPS-Drive: Task-Guided Representation Purification for VLM-based Autonomous Driving}

\author{%
  Jiaxiang Li$^{1}$,
  Yumao Liu$^{1}$,
  Ke Ma$^{1}$\thanks{Corresponding author} \\
  $^{1}$The Hong Kong University of Science and Technology (Guangzhou), Guangzhou, China \\
  \texttt{zmbdsilver@gmail.com},
  \texttt{yliu313@hkust-gz.edu.cn},
  \texttt{kema@hkust-gz.edu.cn}
}

\begin{document}

\maketitle

\begin{abstract}
Vision-Language Models (VLMs) provide a promising foundation for autonomous driving planning, yet bridging semantic reasoning and precise 3D spatial forecasting remains a critical challenge. Existing representation strategies generally follow two paths: text-aligned methods flatten continuous spatial states into symbols, which compromises geometric structure and induces "spatial hallucinations"; dense visual methods preserve spatial topology but overwhelm standard tokenizers with redundant background textures, leading to "representation interference". To address these limitations, we introduce TPS-Drive, a novel framework centered on Task-Guided Representation Purification that empowers VLMs to \emph{\textbf{T}hink in \textbf{P}urified \textbf{S}pace}. At its core, an Agent-Centric Tokenizer utilizes a task-guided vector quantization mechanism supervised by a frozen 3D detection head, which explicitly reallocates limited codebook capacity from pervasive static backgrounds to critical dynamic agents and effectively isolates spatial redundancy. Leveraging this purified spatial vocabulary, TPS-Drive employs a decoupled reasoning pipeline that sequentially performs scene understanding, future forecasting, and action generation. The framework is optimized via a progressive three-stage training paradigm, culminating in reward-driven refinement that surpasses pure imitation learning. Extensive experiments validate our approach: TPS-Drive achieves accurate agent spatial state forecasting and reduces collision rates in open-loop nuScenes evaluations, while establishing new safety records on the rigorous closed-loop NAVSIMv1 and NAVSIMv2 benchmarks.
\end{abstract}

\begin{figure}[!tbp] 
    \centering
    \includegraphics[width=\linewidth]{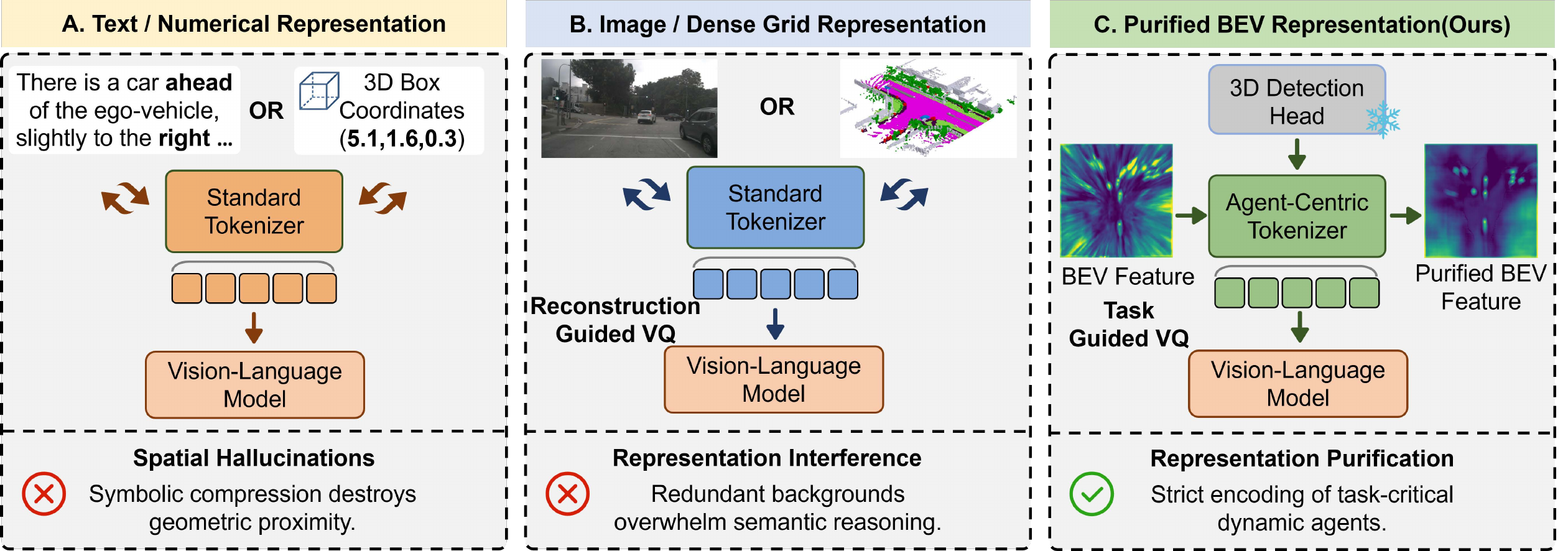} 
    \caption{\textbf{Comparison of representation strategies.} \textbf{(A)} Text and numerical formats induce spatial hallucinations, as symbolic compression destroys geometric proximity. \textbf{(B)} Dense grid representations cause representation interference, where redundant backgrounds overwhelm semantic reasoning. \textbf{(C)} Our \textbf{TPS-Drive} achieves representation purification via task-guided VQ, enabling the strict encoding of task-critical dynamic agents.}
    \label{fig:representation_comparison}
\end{figure}

\section{Introduction}
Vision-Language Models (VLMs) provide a promising foundation for autonomous driving planning through strong semantic reasoning on multimodal inputs\cite{jiang2025survey, zhou2024vision, tian2025large}.
However, their language-centric architectures are not naturally designed to model continuous geometric structures\cite{wang2026vggdrive} and temporal dynamics\cite{zhou2025vlm4d}. This limitation becomes critical in autonomous driving, where executable actions require accurate forecasting of agent spatial state evolution in 3D physical space\cite{guo2024surds}. Thus, designing a representation that bridges the gap between semantic reasoning and precise forecasting of agent spatial states remains a key challenge for VLM-based autonomous driving systems.

Existing methods typically address this challenge through two primary representation strategies, as illustrated in Figure \ref{fig:representation_comparison}. 
The first strategy relies on text\cite{wu2025language,he2024integrating} or numerical\cite{li2025spacedrive} formats (Figure \ref{fig:representation_comparison}A), representing spatial states through either natural language descriptions (e.g., \emph{ahead, right}) or flattened 3D box coordinates. While such text-aligned representations integrate seamlessly with standard VLM tokenizers, mapping continuous spatial states to discrete symbols compromises the underlying geometric structure of physical space\cite{luo2026last}.  Thus, this geometry-agnostic discretization often induces "spatial hallucinations" and limits the VLM's ability to precisely capture the continuous spatial states of dynamic agents.
 
The second strategy employs image or dense grid representations (Figure \ref{fig:representation_comparison}B), such as forecasting future images overlaid with 3D boxes\cite{zeng2025futuresightdrive} or modeling dense 3D occupancy grids\cite{zheng2024occworld}. 
Compared with text-based representations, dense representations better preserve spatial topology. However, they also contain large regions of repetitive or weakly task-relevant content, such as static backgrounds and texture details. Driven by reconstruction-guided Vector Quantization (VQ)\cite{cao2026fastdrivevla}, standard tokenizers in this paradigm treat all dense features equally. 
They disproportionately allocate their limited codebook capacity to static background and redundant textures rather than to critical dynamic agents.
Thus, feeding these spatial tokens into the VLM induces severe "representation interference" that overwhelms the model's semantic reasoning capabilities. 

To overcome these limitations, we introduce TPS-Drive---a framework centered on task-guided representation purification that enables the VLM to \emph{Think in Purified Space}. Rather than flattening continuous 3D coordinates or burdening the model with redundant dense inputs, TPS-Drive constructs an agent-centric, purified Bird’s-Eye-View (BEV) representation (Figure \ref{fig:representation_comparison}C).
At its core, an Agent-Centric Tokenizer employs a task-guided VQ mechanism supervised by a frozen 3D detection head. This design explicitly reallocates limited codebook capacity from static background and redundant textures to critical dynamic agents. 
This "representation purification" effectively isolates spatial redundancy from semantic reasoning, empowering the VLM to ingest purified 3D tokens.

Equipped with this purified spatial vocabulary, TPS-Drive executes a decoupled reasoning pipeline: it first extracts static environmental context and physical rules, then forecasts the spatial state evolution of dynamic agents, and ultimately generates continuous ego trajectories. This pipeline is optimized via a progressive three-stage training paradigm. The process begins with task-guided pretraining for the agent-centric tokenizer. The VLM is then optimized via Supervised Fine-Tuning (SFT) to jointly master scene understanding, future forecasting, and action generation. Ultimately, reward-driven refinement further aligns the final policy with safety-oriented objectives.

Extensive experiments on nuScenes, NAVSIMv1, and NAVSIMv2 validate the effectiveness of TPS-Drive. On nuScenes, TPS-Drive improves agent-centric spatial forecasting and achieves strong collision-avoidance performance in open-loop planning. In closed-loop settings, it delivers robust planning performance on NAVSIMv1 and establishes new safety records under the rigorous interactive pseudo-simulation of NAVSIMv2. The main contributions are as follows:

\begin{itemize}
    \item We propose an Agent-Centric Tokenizer supervised by a frozen 3D detection head. This achieves task-guided representation purification by explicitly isolating spatial redundancy, empowering the VLM to \emph{Think in Purified Space}.
    \item We introduce TPS-Drive, a decoupled framework for scene understanding, future forecasting, and action generation, optimized via a three-stage training paradigm featuring reward-driven refinement that surpasses pure imitation learning.
    \item TPS-Drive achieves exceptional spatial forecasting and minimal collision rates on open-loop nuScenes, while establishing new safety records on the rigorous closed-loop NAVSIMv1 and NAVSIMv2 benchmarks.
\end{itemize}

\section{Related Work}

\subsection{Vision-Language Models in Autonomous Driving}
Recent advances in multimodal learning\cite{zong2024self,shukor2025scaling} have spurred the development of VLMs for end-to-end autonomous driving\cite{yang2023llm4drive,sima2024drivelm}. Within this domain, researchers are increasingly adapting these architectures to process complex traffic scenarios\cite{fruhwirth2025stsbench,brusnicki2026well,zeng2025vision}. For instance, DriveGPT4\cite{xu2024drivegpt4} uses a video-based VLM for textual explanations and control. LMDrive\cite{shao2024lmdrive} processes multimodal inputs alongside language instructions to output closed-loop waypoints\cite{jia2024bench2drive,jia2026bench2drive,dosovitskiy2017carla}. Additionally, DriveMLM\cite{cui2025drivemlm} compresses sensor data via a multimodal tokenizer for behavioral planning alignment. To improve spatial reasoning, OmniDrive\cite{wang2025omnidrive} explicitly aligns 3D perception features with language models. Expanding on this, BEVDriver\cite{winter2025bevdriver} and OccVLA\cite{liu2025occvla} fuse dense Bird’s-Eye-View or occupancy representations directly into the VLM's token space. While these models demonstrate impressive capabilities, they fundamentally face a trade-off between geometric fidelity and computational efficiency\cite{li2026perceptio}, often leading to spatial hallucinations\cite{ogunleye20263d,fan2024hallucination} or representation interference\cite{rongali2026task}. Unlike these approaches, our proposed TPS-Drive introduces a task-guided mechanism to purify spatial features prior to VLM reasoning.

\subsection{World Models for Spatial Forecasting}
World models\cite{liu2026dynflowdrive,tu2025role,zheng2025world4drive} play a critical role in anticipating the spatial state evolution of dynamic agents for safe autonomous planning. Existing methods\cite{feng2025survey,diehl2025dio} typically follow two distinct trajectories. The first leverages dense continuous representations, such as predicting future Bird's-Eye-View maps\cite{fu2026prodrive}, point clouds\cite{yang2024visual,agro2024uno}, or occupancy grids\cite{zheng2024occworld,gu2024dome}. OccWorld \cite{zheng2024occworld} and DriveWorld \cite{min2024driveworld} forecast continuous 3D occupancy volumes to successfully preserve geometric structures. Conversely, driven by the scalability of large language models\cite{wu2024smart,brown2020language}, the second trajectory employs discrete tokenization. Frameworks like GAIA-1 \cite{hu2023gaia}, MILE \cite{hu2022model}, and recent generative world models\cite{bruce2024genie,hafner2023mastering} utilize reconstruction-guided Vector Quantization\cite{esser2021taming} to discretize scenes for autoregressive prediction. However, these standard tokenizers are inherently burdened by representation redundancy\cite{ma2025unitok}, misallocating codebook capacity to massive static backgrounds. To overcome this limitation, TPS-Drive explicitly isolates this spatial redundancy, shifting from global reconstruction to task-guided purification to yield highly distilled 3D tokens focused entirely on dynamic agents.

\subsection{Spatial Representation and Scene Tokenization}
3D spatial representation\cite{chen2024spatialvlm} in VLMs typically follows text-based serialization\cite{hua2026unleashing,yuan2026boosting} or dense representation paradigms\cite{halacheva2025gaussianvlm,cheng2024spatialrgpt}. Text-based methods like LMDrive\cite{shao2024lmdrive} and Drivegpt4\cite{xu2024drivegpt4} discretize 3D coordinates into numerical tokens, which often compromises the underlying geometric structure. Conversely, dense tokenizers like VQ-VAE\cite{van2017neural} rely on reconstruction objectives that overwhelm the codebook with redundant background textures. To address this, OccLLaMA\cite{wei2024occllama} discretizes semantic occupancy to explicitly handle spatial sparsity. Furthering this trend, TPS-Drive employs an agent-centric tokenizer to yield distilled 3D tokens focused exclusively on dynamic agents, empowering the model to think in purified space.

\begin{figure*}[t]
    \centering
    \includegraphics[width=1.0\linewidth]{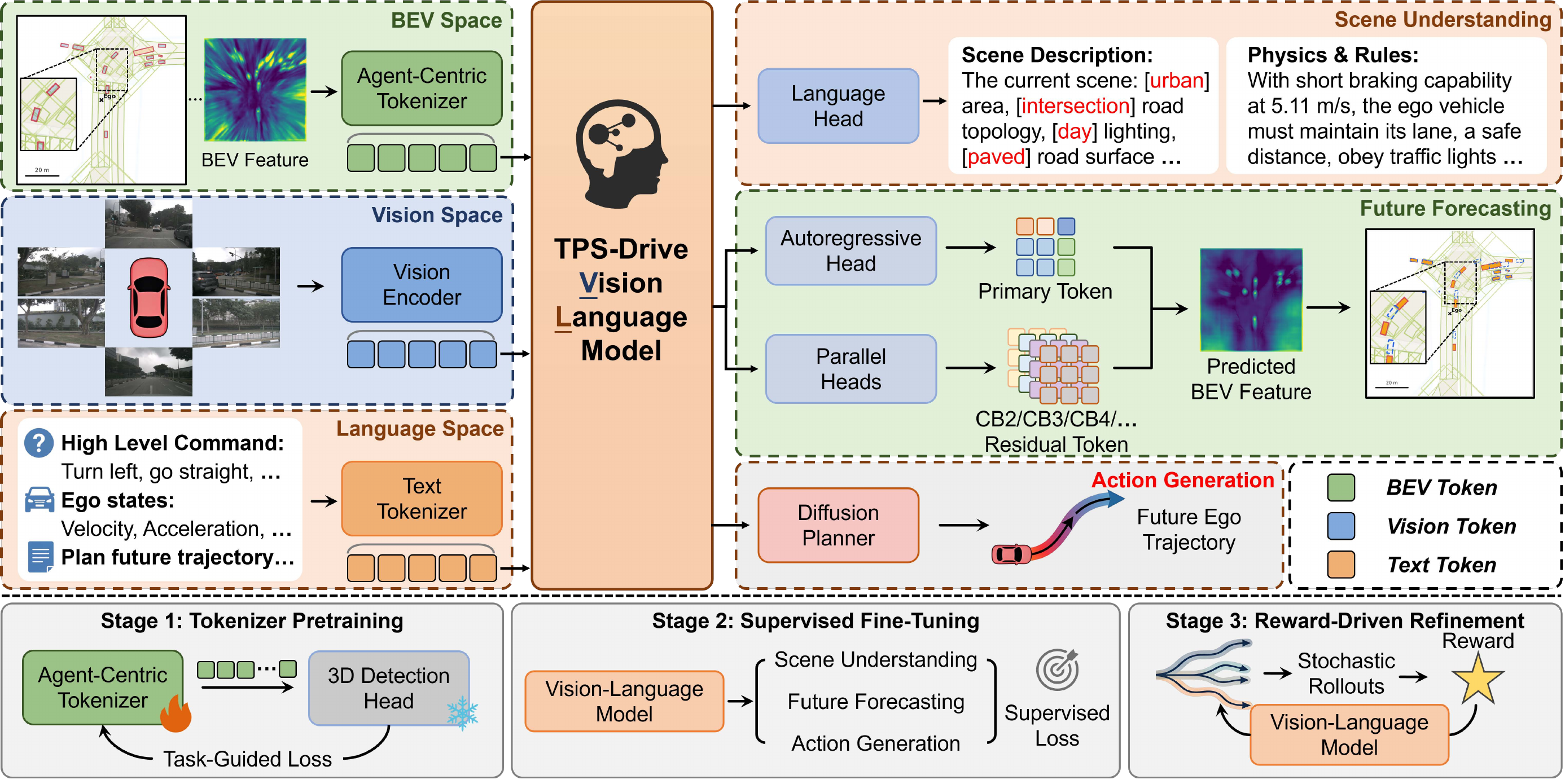} 
    \caption{\textbf{Overview of TPS-Drive.} The decoupled VLM explicitly separates scene understanding, future forecasting, and action generation. The bottom panel illustrates the progressive three-stage training paradigm: tokenizer pretraining, supervised fine-tuning, and reward-driven refinement.}
    \label{fig:method_overview}
\end{figure*}

\section{Method}

\subsection{Preliminary}
As illustrated in Figure~\ref{fig:method_overview}, TPS-Drive formulates autonomous driving as a decoupled three-stage reasoning process. At time $t$, the ego-vehicle processes multi-view images $\mathcal{I}_t = \{\mathcal{I}_t^c\}_{c=1}^{C}$ from $C$ surrounding cameras and an ego-centric Bird's-Eye-View (BEV) feature map $\mathcal{B}_t \in \mathbb{R}^{D \times H \times W}$\cite{li2024bevformer}, where $D$, $H$, and $W$ denote the feature dimension and spatial resolutions. Alongside this visual context, the system receives a high-level navigation command $\mathbf{c}$ (e.g., \emph{turn left}), a task prompt $\mathbf{p}$, and current vehicle kinematics $\mathbf{k}_t = [\mathbf{v}_t, \mathbf{a}_t]$ comprising velocity and acceleration. The ultimate objective is to output a dynamically feasible future trajectory $\boldsymbol{\tau}_{t+1:t+T_{\text{plan}}}$ over a planning horizon $T_{\text{plan}}$.

To effectively bridge these multimodal inputs and the continuous trajectory, we introduce an explicit intermediate representation. We construct a discrete spatial vocabulary $\mathcal{V}$ via an agent-centric tokenizer (Sec.~\ref{sec:tokenization}). Guided by a frozen 3D detection head, this vocabulary encodes the future BEV feature $\mathcal{B}_{t+\Delta t}$ into a distilled token sequence $\mathbf{z}_{t+\Delta t} \in \mathcal{V}^{N}$, where $N$ denotes the sequence length. This tokenization explicitly isolates spatial redundancy while preserving the state of dynamic agents.

Grounded in this purified representation, the three-stage process operates sequentially:
(1) \textbf{Scene Understanding:} A vision-language backbone first extracts $\mathbf{s}_t$, a unified structured representation encapsulating both the scene description and underlying physics rules from the current context.
(2) \textbf{Future Forecasting:} Conditioned on $\mathbf{s}_t$ and the current observations, the model predicts the future world tokens $\mathbf{z}_{t+\Delta t}$ to anticipate spatial evolution.
(3) \textbf{Action Generation:} Finally, a conditional diffusion planner derives the continuous trajectory $\boldsymbol{\tau}_{t+1:t+T_{\text{plan}}}$ based on these predicted states.

\subsection{Agent-Centric Tokenization}
\label{sec:tokenization}
\textbf{Task-Guided Primary Codebook.} The foundation of our spatial vocabulary is a primary codebook $\mathcal{V}_1$, designed to extract the spatial states of dynamic agents from BEV features into discrete tokens. An encoder $E_\phi$ projects a BEV feature map $\mathcal{B}$ into a lower-resolution latent grid $\mathbf{e} \in \mathbb{R}^{D' \times H' \times W'}$, where $D'$ is the latent feature dimension and $H' \times W' = N$ defines the downsampled spatial resolution. Each spatial position $\mathbf{e}_{i,j} \in \mathbb{R}^{D'}$ is quantized to its nearest entry in $\mathcal{V}_1$, producing a discrete token map $\mathbf{z} \in \mathcal{V}_1^{N}$:
\begin{equation}
    z_{i,j} = \arg\min_{m} \bigl\| \mathbf{e}_{i,j} - \mathbf{v}_m \bigr\|_2, \qquad \mathbf{q}_{i,j} = \mathbf{v}_{z_{i,j}},
    \label{eq:vq}
\end{equation}
where $\mathbf{v}_m \in \mathbb{R}^{D'}$ is a codebook entry indexed by $m$, and $\mathbf{q} \in \mathbb{R}^{D' \times H' \times W'}$ is the quantized representation. A decoder $D_\psi$ then reconstructs the purified feature map as $\hat{\mathcal{B}} = D_\psi(\mathbf{q})$.

To prevent this codebook from passively encoding task-irrelevant static background, we utilize a frozen, pretrained CenterPoint3D detection head\cite{yin2021center,zhou2018voxelnet} for explicit supervision. The training objective balances reconstruction accuracy, task relevance, and codebook stability:
\begin{equation}
    \mathcal{L}_{\text{primary}} 
    = \underbrace{\lambda_r \, \mathcal{L}_{\text{rec}}(\hat{\mathcal{B}}, \mathcal{B})}_{\text{reconstruction}} 
    + \underbrace{\lambda_h \, \mathcal{L}_{\text{hm}} + \lambda_b \, \mathcal{L}_{\text{box}}}_{\text{task guidance}} 
    + \underbrace{\mathcal{L}_{\text{vq}} + \beta \, \mathcal{L}_{\text{commit}}}_{\text{codebook learning}},
    \label{eq:primary_loss}
\end{equation}
where $\mathcal{L}_{\text{hm}}$ and $\mathcal{L}_{\text{box}}$ denote the heatmap and 3D bounding box losses, respectively, and $\lambda_r, \lambda_h, \lambda_b, \beta$ are balancing weights. The terms $\mathcal{L}_{\text{vq}}$ and $\mathcal{L}_{\text{commit}}$ represent standard VQ-VAE objectives~\cite{van2017neural}. While our primary objective is representation purification, the weighted reconstruction loss $\lambda_r \, \mathcal{L}_{\text{rec}}$ remains essential: although it may inadvertently preserve minor background details, it prevents the purified features from structurally deteriorating or deviating excessively from the original spatial distribution. This formulation compels $\mathcal{V}_1$ to learn a concise vocabulary of agent-centric patterns, enabling the vision-language backbone to effectively process spatial dynamics.

\textbf{Residual Refinement Layers.} Although the task-guided primary codebook captures core structural patterns, a single quantization level inherently loses precise positional nuances. To recover these details, we incorporate $L-1$ supplementary codebooks, $\{\mathcal{V}_\ell\}_{\ell=2}^{L}$, arranged in a residual hierarchy. Each subsequent codebook quantizes the residual error from prior layers, producing $L$ token maps $\{\mathbf{z}^{(\ell)}\}_{\ell=1}^{L}$ with corresponding quantized representations $\{\mathbf{q}^{(\ell)}\}_{\ell=1}^{L}$ that depict the purified BEV features at increasingly finer resolutions. The final high-fidelity reconstruction is obtained by decoding the cumulative quantized features: $\tilde{\mathcal{B}} = D_\psi\bigl(\sum_{\ell=1}^{L} \mathbf{q}^{(\ell)}\bigr)$.

We refine this architecture through a two-phase optimization schedule. After training the primary codebook $\mathcal{V}_1$, we freeze its weights and proceed to optimize the residual codebooks. Since $\mathcal{V}_1$ establishes a semantically comprehensive structural foundation, only the primary tokens $\mathbf{z}^{(1)}$ are input to the vision-language backbone as an autoregressive prediction target. The residual tokens are predicted in parallel using lightweight classification heads. This hierarchical structure keeps the autoregressive sequence compact for computational efficiency, while the residual layers retain high-fidelity spatial details.

\subsection{Decoupled Reasoning Architecture}
\label{sec:reasoning}
\textbf{Scene Understanding.} The first stage distills the driving environment into a unified structured representation $\mathbf{s}_t$, which encapsulates both a template-constrained scene description and underlying physics rules. Using a predefined vocabulary, this representation categorizes essential environmental attributes (e.g., \emph{area type, road topology, lighting}) alongside explicit physical constraints (e.g., \emph{short braking capability}). To obtain reliable supervision targets, we annotate the training data using the Qwen3.5-27B\cite{team2026qwen3} model to generate the structured representations from raw sensor inputs and contextual metadata, followed by manual inspection to ensure correctness and consistency\cite{liu2025traffic}. The resulting outputs are highly consistent, easily parsed, and supervised via a cross-entropy loss $\mathcal{L}_{\text{scene}}$ against these curated annotations. Crucially, because $\mathbf{s}_t$ is generated prior to the future world tokens $\mathbf{z}_{t+\Delta t}$, it provides a comprehensive structured prior that conditions the subsequent spatial predictions.

\textbf{Future Forecasting.} The second stage forecasts the future spatial state within the discrete token space. Let $\mathbf{z}_{t+\Delta t}^{(1)}$ denote the primary tokens of the future purified BEV map at time $t + \Delta t$, and let $\{\mathbf{z}_{t+\Delta t}^{(\ell)}\}_{\ell=2}^{L}$ denote the corresponding residual tokens. The complete conditioning context is aggregated as $\mathbf{x}_t = \{\mathcal{I}_t, \mathcal{B}_t, \mathbf{c}, \mathbf{p}, \mathbf{k}_t, \mathbf{s}_t\}$. The vision-language backbone autoregressively predicts the primary token sequence in raster order\cite{dutta2023implementation} by modeling $p\bigl(z_{t+\Delta t, n}^{(1)} \,\big|\, \mathbf{x}_t, \mathbf{z}_{t+\Delta t, <n}^{(1)}\bigr)$. Concurrently, $L-1$ classification heads $\{g_\ell\}_{\ell=2}^{L}$ predict the residual tokens based on the hidden states $\mathbf{h}_{t+\Delta t}$ produced by the backbone during primary token generation. The overall objective combines autoregressive primary prediction with parallel residual classification:
\begin{equation}
    \mathcal{L}_{\text{world}} 
    = \underbrace{-\sum_{n=1}^{N} \log p\bigl(z_{t+\Delta t, n}^{(1)} \,\big|\, \mathbf{x}_t, \mathbf{z}_{t+\Delta t, <n}^{(1)}\bigr)}_{\text{autoregressive primary}}
    \;+\;
    \underbrace{\lambda_{\text{res}} \sum_{\ell=2}^{L} \mathcal{L}_{\text{CE}}\bigl(g_\ell(\mathbf{h}_{t+\Delta t}),\, \mathbf{z}_{t+\Delta t}^{(\ell)}\bigr)}_{\text{parallel residual}},
    \label{eq:world_loss}
\end{equation}
where $\lambda_{\text{res}}$ is a balancing weight and $\mathcal{L}_{\text{CE}}$ denotes the standard cross-entropy loss.

\textbf{Action Generation.} The final stage translates the perceived environment and the anticipated agent positions into a continuous control trajectory. To effectively model the multi-modal distribution of human driving behaviors, we utilize a conditional diffusion model~\cite{zhang2023adding,kang2023efficient}. The planner is conditioned on a unified feature vector $\mathbf{f}_t$, which concatenates the hidden states of the predicted future tokens $\mathbf{h}_{t+\Delta t}$, the structured representation $\mathbf{s}_t$, and an action start prompt $\mathbf{p}_{\text{start}}$. A denoising Transformer $\epsilon_\theta$ is trained to estimate the noise injected during the forward diffusion process\cite{ho2020denoising} by minimizing:
\begin{equation}
    \mathcal{L}_{\text{plan}} = \mathbb{E}_{k, \boldsymbol{\tau}^{0}, \boldsymbol{\epsilon}} \Bigl[ \bigl\| \boldsymbol{\epsilon} - \epsilon_\theta(\boldsymbol{\tau}^{k}, k, \mathbf{f}_t) \bigr\|_2^2 \Bigr],
    \label{eq:diffusion_loss}
\end{equation}
where $k \in \{1, \ldots, K\}$ denotes the diffusion timestep, $\boldsymbol{\tau}^{0}$ is the ground-truth trajectory, $\boldsymbol{\epsilon} \sim \mathcal{N}(\mathbf{0}, \mathbf{I})$ is the standard Gaussian noise sampled during the forward diffusion process, and $\boldsymbol{\tau}^{k}$ represents the noised trajectory at step $k$. Optimizing the diffusion-based planner independently of the vision-language backbone significantly stabilizes training.

\begin{table*}[t]
\centering
\renewcommand{\arraystretch}{1.05}
\setlength{\tabcolsep}{2.0pt}
\footnotesize
\caption{End-to-end trajectory planning experiments on nuScenes. * indicates that the ego status is additionally used. \textbf{Bold} and \underline{underline} indicate the best and second-best VLM results. All average columns are highlighted in gray.}
\label{tab:neurips_merged_final}
\resizebox{\textwidth}{!}{
\begin{tabular}{l cccc cccc cccc cccc}
\toprule
 & \multicolumn{8}{c}{\textbf{ST-P3 metrics}} & \multicolumn{8}{c}{\textbf{UniAD metrics}} \\
\cmidrule(lr){2-9} \cmidrule(lr){10-17}
\multicolumn{1}{c}{\textbf{Method}} & \multicolumn{4}{c}{\textbf{L2 (m)} $\downarrow$} & \multicolumn{4}{c}{\textbf{CR (\%)} $\downarrow$} & \multicolumn{4}{c}{\textbf{L2 (m)} $\downarrow$} & \multicolumn{4}{c}{\textbf{CR (\%)} $\downarrow$} \\
\cmidrule(lr){2-5} \cmidrule(lr){6-9} \cmidrule(lr){10-13} \cmidrule(lr){14-17}
 & 1s & 2s & 3s & \cellcolor{gray!15}\textbf{Avg.} & 1s & 2s & 3s & \cellcolor{gray!15}\textbf{Avg.} & 1s & 2s & 3s & \cellcolor{gray!15}\textbf{Avg.} & 1s & 2s & 3s & \cellcolor{gray!15}\textbf{Avg.} \\
\midrule
\multicolumn{17}{c}{\textit{E2E-based Methods}} \\
\midrule
ST-P3*\cite{hu2022st} {\scriptsize [ECCV'22]}         & 1.33 & 2.11 & 2.90 & \cellcolor{gray!15}2.11 & 0.23 & 0.62 & 1.27 & \cellcolor{gray!15}0.71 & - & - & - & \cellcolor{gray!15}- & - & - & - & \cellcolor{gray!15}- \\
UniAD*\cite{hu2023planning} {\scriptsize [CVPR'23]}          & - & - & - & \cellcolor{gray!15}- & - & - & - & \cellcolor{gray!15}- & 0.20 & 0.42 & 0.75 & \cellcolor{gray!15}0.46 & 0.02 & 0.25 & 0.84 & \cellcolor{gray!15}0.37 \\
BEV-Planner\cite{li2024ego} {\scriptsize [CVPR'24]}     & 0.30 & 0.52 & 0.83 & \cellcolor{gray!15}0.55 & 0.10 & 0.37 & 1.30 & \cellcolor{gray!15}0.59 & - & - & - & \cellcolor{gray!15}- & - & - & - & \cellcolor{gray!15}- \\
BEV-Planner*\cite{li2024ego} {\scriptsize [CVPR'24]}    & 0.16 & 0.32 & 0.57 & \cellcolor{gray!15}0.35 & 0.00 & 0.29 & 0.73 & \cellcolor{gray!15}0.34 & - & - & - & \cellcolor{gray!15}- & - & - & - & \cellcolor{gray!15}- \\
MomAD\cite{song2025don} {\scriptsize [CVPR'25]}           & 0.31 & 0.57 & 0.91 & \cellcolor{gray!15}0.60 & 0.01 & 0.05 & 0.22 & \cellcolor{gray!15}0.09 & 0.43 & 0.88 & 1.62 & \cellcolor{gray!15}0.98 & 0.06 & 0.16 & 0.68 & \cellcolor{gray!15}0.30 \\
HE-Drive\cite{wang2024he} {\scriptsize [ICLR'25]}        & 0.31 & 0.58 & 0.93 & \cellcolor{gray!15}0.61 & 0.01 & 0.05 & 0.16 & \cellcolor{gray!15}0.07 & - & - & - & \cellcolor{gray!15}- & - & - & - & \cellcolor{gray!15}- \\
DiffusionDrive\cite{liao2025diffusiondrive} {\scriptsize [CVPR'25]}  & 0.31 & 0.62 & 1.03 & \cellcolor{gray!15}0.65 & 0.03 & 0.06 & 0.19 & \cellcolor{gray!15}0.09 & 0.43 & 1.01 & 1.83 & \cellcolor{gray!15}1.09 & 0.05 & 0.14 & 0.55 & \cellcolor{gray!15}0.25 \\
PRIX\cite{wozniak2026prix} {\scriptsize [RAL'26]}             & 0.26 & 0.53 & 0.93 & \cellcolor{gray!15}0.57 & 0.00 & 0.04 & 0.18 & \cellcolor{gray!15}0.07 & - & - & - & \cellcolor{gray!15}- & - & - & - & \cellcolor{gray!15}- \\
\midrule
\multicolumn{17}{c}{\textit{VLM-based or World Model-based Methods}} \\
\midrule
RDA-Driver*\cite{huang2024making} {\scriptsize [ECCV'24]}     & 0.17 & 0.37 & 0.69 & \cellcolor{gray!15}0.41 & \underline{0.01} & \textbf{0.05} & 0.26 & \cellcolor{gray!15}\underline{0.11} & \underline{0.23} & 0.73 & 1.54 & \cellcolor{gray!15}0.83 & \textbf{0.00} & 0.13 & 0.83 & \cellcolor{gray!15}0.32 \\
Doe-1\cite{zheng2024doe} {\scriptsize [arXiv'24]}          & 0.37 & 0.67 & 1.07 & \cellcolor{gray!15}0.70 & 0.02 & 0.14 & 0.47 & \cellcolor{gray!15}0.21 & 0.50 & 1.18 & 2.11 & \cellcolor{gray!15}1.26 & 0.04 & 0.37 & 1.19 & \cellcolor{gray!15}0.53 \\
OccWorld\cite{zheng2024occworld} {\scriptsize [ECCV'24]}        & 0.39 & 0.73 & 1.18 & \cellcolor{gray!15}0.77 & 0.11 & 0.19 & 0.67 & \cellcolor{gray!15}0.32 & 0.52 & 1.27 & 2.41 & \cellcolor{gray!15}1.40 & 0.12 & 0.40 & 2.08 & \cellcolor{gray!15}0.87 \\
OmniDrive\cite{wang2025omnidrive} {\scriptsize [CVPR'25]}       & 0.40 & 0.80 & 1.32 & \cellcolor{gray!15}0.84 & 0.04 & 0.46 & 2.32 & \cellcolor{gray!15}0.94 & 0.54 & 1.23 & 2.53 & \cellcolor{gray!15}1.43 & 0.10 & 0.43 & 1.51 & \cellcolor{gray!15}0.68 \\
OmniDrive*\cite{wang2025omnidrive} {\scriptsize [CVPR'25]}      & \textbf{0.14} & \underline{0.29} & 0.55 & \cellcolor{gray!15}0.33 & \textbf{0.00} & 0.13 & 0.78 & \cellcolor{gray!15}0.30 & 0.25 & 0.49 & 1.11 & \cellcolor{gray!15}0.62 & \underline{0.02} & 0.13 & 0.65 & \cellcolor{gray!15}0.27 \\
EMMA*\cite{hwang2024emma} {\scriptsize [TMLR'25]}           & \textbf{0.14} & \underline{0.29} & 0.54 & \cellcolor{gray!15}\underline{0.32} & 0.03 & 0.08 & \underline{0.25} & \cellcolor{gray!15}0.12 & - & - & - & \cellcolor{gray!15}- & - & - & - & \cellcolor{gray!15}- \\
FSDrive\cite{zeng2025futuresightdrive} {\scriptsize [NeurIPS'25]}      & 0.28 & 0.52 & 0.80 & \cellcolor{gray!15}0.53 & 0.06 & 0.13 & 0.32 & \cellcolor{gray!15}0.17 & 0.40 & 0.89 & 1.60 & \cellcolor{gray!15}0.96 & 0.07 & 0.12 & 1.02 & \cellcolor{gray!15}0.40 \\
FSDrive*\cite{zeng2025futuresightdrive} {\scriptsize [NeurIPS'25]}     & \textbf{0.14} & \textbf{0.25} & \textbf{0.46} & \cellcolor{gray!15}\textbf{0.28} & 0.03 & \underline{0.06} & \textbf{0.21} & \cellcolor{gray!15}\textbf{0.10} & \textbf{0.18} & \underline{0.39} & \textbf{0.77} & \cellcolor{gray!15}\textbf{0.45} & \textbf{0.00} & \underline{0.06} & \underline{0.42} & \cellcolor{gray!15}\underline{0.16} \\
\midrule
\textbf{TPS-Drive (Ours)}               & 0.32 & 0.53 & 0.81 & \cellcolor{gray!15}0.55 & \underline{0.01} & 0.13 & 0.43 & \cellcolor{gray!15}0.19 & 0.42 & 0.88 & 1.52 & \cellcolor{gray!15}0.94 & \underline{0.02} & 0.17 & 1.19 & \cellcolor{gray!15}0.46 \\
\textbf{TPS-Drive* (Ours)}              & \underline{0.15} & \underline{0.29} & \underline{0.53} & \cellcolor{gray!15}\underline{0.32} & \textbf{0.00} & \textbf{0.05} & \underline{0.25} & \cellcolor{gray!15}\textbf{0.10} & \underline{0.23} & \textbf{0.38} & \underline{0.85} & \cellcolor{gray!15}\underline{0.49} & \textbf{0.00} & \textbf{0.04} & \textbf{0.38} & \cellcolor{gray!15}\textbf{0.14} \\
\bottomrule
\end{tabular}
}
\end{table*}

\subsection{Training Strategy}
\label{sec:training}
We adopt a progressive three-stage training strategy that accommodates the distinct optimization dynamics of the tokenizer, vision-language backbone, and diffusion planner.

\textbf{Stage 1: Tokenizer Pretraining.} The spatial tokenizer is trained in two phases. First, the primary encoder $E_\phi$, codebook $\mathcal{V}_1$, and decoder $D_\psi$ are optimized under the task-guided objective $\mathcal{L}_{\text{primary}}$. These components are then frozen while the residual codebooks $\{\mathcal{V}_\ell\}_{\ell=2}^{L}$ are trained under the reconstruction objective, yielding a universal and robust tokenizer.

\textbf{Stage 2: Supervised Fine-Tuning.} We adapt the pretrained vision-language backbone through supervised fine-tuning. The training objective jointly optimizes the three reasoning stages by combining $\mathcal{L}_{\text{scene}}$ for structured scene understanding, $\mathcal{L}_{\text{world}}$ (Eq.~\ref{eq:world_loss}) for future world modeling, and $\mathcal{L}_{\text{plan}}$ (Eq.~\ref{eq:diffusion_loss}) for trajectory generation. The vision encoder and language model are fine-tuned efficiently, while the specialized projection, classification, and diffusion heads are fully updated.

\textbf{Stage 3: Reward-Driven Refinement.} To refine driving behavior beyond supervised imitation, we apply an end-to-end grouped relative optimization strategy\cite{shao2024deepseekmath}. For each sample, the policy generates multiple stochastic rollouts containing both predicted world tokens $\mathbf{z}_{t+\Delta t}$ and diffusion-generated trajectories $\boldsymbol{\tau}$, which are scored by an offline reward based on geometric errors and motion smoothness. Rewards are normalized within each group to form relative advantages, eliminating the need for a separate value network. The world-modeling branch is updated via an advantage-weighted policy objective, while the diffusion planner is refined through a reward-weighted denoising loss.
\section{Experiments}
\label{sec:experiments}

\begin{table*}[t]
\centering
\renewcommand{\arraystretch}{1.1}
\setlength{\tabcolsep}{6pt} 
\footnotesize
\caption{Comparison on NAVSIMv1 with PDMS.}
\label{tab:navsim_final}
\begin{tabular}{l c ccccc >{\columncolor{gray!15}}c}
\toprule
\multicolumn{1}{c}{\textbf{Method}} & \textbf{Sensors} & \textbf{NC}$\uparrow$ & \textbf{DAC}$\uparrow$ & \textbf{TTC}$\uparrow$ & \textbf{C}$\uparrow$ & \textbf{EP}$\uparrow$ & \textbf{PDMS}$\uparrow$ \\
\midrule
Human & - & 100.0 & 100.0 & 100.0 & 99.9 & 87.5 & 94.8 \\
\midrule
DiffusionDrive\cite{liao2025diffusiondrive} {\scriptsize [CVPR'25]} & 3$\times$C + L & 98.2 & 96.2 & 94.7 & \textbf{100.0} & 82.2 & 88.1 \\
DrivingGPT\cite{chen2025drivinggpt} {\scriptsize [ICCV'25]} & 1$\times$C & \textbf{98.9} & 90.7 & 94.9 & 95.6 & 79.7 & 82.4 \\
WoTE\cite{li2025end} {\scriptsize [ICCV'25]} & 3$\times$C + L & 98.5 & 96.8 & 94.4 & 99.9 & 81.9 & 88.3 \\
AutoVLA\cite{zhou2025autovla} {\scriptsize [NeurIPS'25]} & 3$\times$C & 98.4 & 95.6 & \textbf{98.0} & 99.9 & 81.9 & 89.1 \\
Recogdrive\cite{li2025recogdrive} {\scriptsize [ICLR'26]} & 3$\times$C & 98.2 & \textbf{97.8} & 95.2 & 99.8 & \textbf{83.5} & 89.6 \\
\midrule
\textbf{TPS-Drive (Ours)} & 3$\times$C & 98.7 & 97.0 & 96.3 & \textbf{100.0} & 82.7 & \textbf{89.7} \\
\bottomrule
\end{tabular}
\end{table*}
\begin{table*}[t]
\centering
\renewcommand{\arraystretch}{1.1}
\setlength{\tabcolsep}{5.2pt} 
\footnotesize
\caption{Comparison on NAVSIMv2 with EPDMS.}
\label{tab:navsimv2_epdms_updated}
\begin{tabular}{l ccccc cccc >{\columncolor{gray!15}}c}
\toprule
\multicolumn{1}{c}{\textbf{Method}} & \textbf{NC}$\uparrow$ & \textbf{DAC}$\uparrow$ & \textbf{DDC}$\uparrow$ & \textbf{TLC}$\uparrow$ & \textbf{EP}$\uparrow$ & \textbf{TTC}$\uparrow$ & \textbf{LK}$\uparrow$ & \textbf{HC}$\uparrow$ & \textbf{EC}$\uparrow$ & \textbf{EPDMS}$\uparrow$ \\
\midrule
TransFuser\cite{chitta2022transfuser} {\scriptsize [TPAMI'22]}      & 96.9 & 89.9 & 97.8 & 99.7 & 87.1 & 95.4 & 92.7 & \textbf{98.3} & 87.2 & 76.7 \\
Ego Status\cite{li2024ego} {\scriptsize [CVPR'24]}       & 93.1 & 77.9 & 92.7 & 99.6 & 86.0 & 91.5 & 89.4 & \textbf{98.3} & 85.4 & 64.0 \\
DiffusionDrive\cite{liao2025diffusiondrive} {\scriptsize [CVPR'25]}   & 98.2 & 95.9 & 99.4 & \textbf{99.8} & 87.5 & 97.3 & 96.8 & \textbf{98.3} & \textbf{87.7} & 84.5 \\
HydraMDP++\cite{li2025hydra} {\scriptsize [CVPR'25]}       & \textbf{98.5} & 98.5 & 99.5 & 99.7 & 87.4 & 97.9 & 95.8 & 98.2 & 75.7 & 85.6 \\
DriveVLA-W0\cite{li2025drivevla} {\scriptsize [CVPR'25]}      & \textbf{98.5} & \textbf{99.1} & 98.0 & 99.7 & 86.4 & \textbf{98.1} & 93.2 & 97.9 & 58.9 & 86.1 \\
\midrule
\textbf{TPS-Drive (Ours)} & \textbf{98.5} & 98.6 & \textbf{99.7} & \textbf{99.8} & \textbf{87.6} & 98.0 & \textbf{97.1} & \textbf{98.3} & 82.5 & \textbf{86.7} \\
\bottomrule
\end{tabular}
\end{table*}
\begin{table*}[t]
\centering
\renewcommand{\arraystretch}{1.15}
\setlength{\tabcolsep}{6pt}
\footnotesize
\caption{Evaluation of agent spatial forecasting on nuScenes.}
\label{tab:future_3d_detection_tps}
\begin{tabular}{l >{\columncolor{gray!15}}c cccccc}
\toprule
\textbf{Method} & \textbf{NDS (\%)} $\uparrow$ & \textbf{mAP (\%)} $\uparrow$ & \textbf{mATE} $\downarrow$ & \textbf{mASE} $\downarrow$ & \textbf{mAOE} $\downarrow$ & \textbf{mAVE} $\downarrow$ & \textbf{mAAE} $\downarrow$ \\
\midrule
BEVWorld\cite{zhang2024bevworld} {\scriptsize [arXiv'24]}      & 25.72 & 21.03 & 0.647 & 0.526 & 0.848 & 1.179 & 0.461 \\
Drive-OccWorld\cite{yang2025driving} {\scriptsize [AAAI'25]} & 20.80 & 20.20 & 0.739 & 0.761 & 0.865 & 1.069 & 0.565 \\
WoTE\cite{li2025end} {\scriptsize [ICCV'25]}           & 30.01 & 22.77 & 0.653 & 0.534 & 0.757 & 0.839 & 0.388 \\
\midrule
\textbf{TPS-Drive (Ours)}              & \textbf{34.60} & \textbf{24.03} & \textbf{0.610} & \textbf{0.421} & \textbf{0.704} & \textbf{0.725} & \textbf{0.300} \\
\bottomrule
\end{tabular}
\end{table*}

\begin{table}[t]
\centering
\renewcommand{\arraystretch}{1.15}
\setlength{\tabcolsep}{4.5pt}
\footnotesize
\caption{Ablation studies on the core components of TPS-Drive.}
\label{tab:ablation_components}
\begin{tabular}{ccc c ccc}
\toprule
\multicolumn{3}{c}{\textbf{TPS-Drive Components}} & \textbf{World Model} & \multicolumn{3}{c}{\textbf{Planning Metrics}} \\
\cmidrule(lr){1-3} \cmidrule(lr){4-4} \cmidrule(lr){5-7}
Task-Guided & Residual & Reward-Driven & \multirow{2}{*}{\textbf{NDS (\%)} $\uparrow$} & \multirow{2}{*}{\textbf{L2 (m)} $\downarrow$} & \multirow{2}{*}{\textbf{CR (\%)} $\downarrow$} & \multirow{2}{*}{\textbf{EPDMS} $\uparrow$} \\
Tokenizer & Layers ($L=4$) & Refinement & & & & \\
\midrule
 & & & 17.31 & 0.64 & 0.33 & 81.1 \\
\checkmark & & & 30.03 & 0.63 & 0.28 & 83.7 \\
\checkmark & \checkmark & & 33.78 & 0.61 & 0.26 & 84.1 \\
\rowcolor{gray!15} 
\checkmark & \checkmark & \checkmark & \textbf{34.60} & \textbf{0.55} & \textbf{0.19} & \textbf{86.7} \\
\bottomrule
\end{tabular}
\end{table}

\subsection{Experimental Setup}
\label{sec:setup}

\paragraph{Datasets and Metrics.} 
We evaluated TPS-Drive across three prominent autonomous driving benchmarks: nuScenes \cite{caesar2020nuscenes}, NAVSIMv1 \cite{dauner2024navsim}, and NAVSIMv2 \cite{cao2025pseudo}. For spatial forecasting and open-loop planning, we utilized the nuScenes dataset. Trajectory planning was assessed using L2 distance and Collision Rate (CR)\cite{hu2022st, hu2023planning}, while future world modeling was measured via the NuScenes Detection Score (NDS) and mean Average Precision (mAP) from agent-centric forecasts. To evaluate closed-loop interactive driving, we employed NAVSIMv1 and NAVSIMv2. NAVSIMv1 performance is captured by the Predictive Driver Model Score (PDMS), whereas NAVSIMv2 uses the Extended EPDMS to enforce stricter regulatory metrics, such as traffic light compliance.

\paragraph{Implementation Details.} 
Our framework is built on the Qwen3.5-VL-2B\cite{team2026qwen3} foundation model and trained using a progressive three-stage pipeline. The BEV encoder first compresses $200\times200$ spatial features to a $25\times25$ resolution. To discretize this space, we employ a primary codebook of 8,192 entries alongside $L=4$ residual codebooks designed to capture fine-grained geometric details. Training begins with task-guided tokenizer pretraining, progresses to supervised fine-tuning of the decoupled architecture, and concludes with a stage of reward-driven refinement. This final stage optimizes the driving policy using simulated rewards to align the model with strict safety regulations.

\subsection{Main Results}
\label{sec:main_results}
\paragraph{Open-Loop Planning (nuScenes).} 
As summarized in Table \ref{tab:neurips_merged_final}, TPS-Drive establishes a new standard for collision avoidance among VLM-based autonomous driving models. The base model achieved a 0.19\% CR under the ST-P3\cite{hu2022st} protocol, substantially outperforming baselines such as OmniDrive\cite{wang2025omnidrive} (0.94\% CR) and OccWorld\cite{zheng2024occworld} (0.32\% CR). When incorporating ego-status, TPS-Drive$^*$ achieved a record-low CR of 0.10\% (ST-P3) and 0.14\% (UniAD), effectively surpassing leading systems like FSDrive$^*$\cite{zeng2025futuresightdrive}. These results validate that cleanly decoupling task-guided spatial understanding from continuous control effectively bridges the gap between high-level semantic reasoning and safe, precise maneuver generation.

\paragraph{Closed-Loop Planning (NAVSIM).} 
The framework also demonstrated exceptional robustness in interactive, closed-loop environments. On NAVSIMv1 (Table \ref{tab:navsim_final}), TPS-Drive achieved a PDMS of 89.7, surpassing recent top-performing models such as Recogdrive\cite{li2025recogdrive} (89.6) and AutoVLA\cite{zhou2025autovla} (89.1). On the more rigorous NAVSIMv2 platform (Table \ref{tab:navsimv2_epdms_updated}), it attained an EPDMS of 86.7, outperforming both concurrent VLA methods like DriveVLA-W0\cite{li2025drivevla} (86.1) and expert-level planners such as HydraMDP++\cite{li2025hydra} (85.6). Notably, the model maintained high scores across granular regulatory metrics, including 99.8\% for traffic light compliance and 97.1\% for lane keeping.

\paragraph{Agent-Centric Spatial Reasoning.} 
To verify the depth of our model's spatial reasoning, we evaluated its agent-centric forecasting accuracy on nuScenes (Table \ref{tab:future_3d_detection_tps}). By feeding predicted BEV features into a fine-tuned detection head, TPS-Drive achieved 34.60\% NDS and 24.03\% mAP. This performance significantly improves upon established world models such as WoTE\cite{li2025end} (30.01\% NDS) and BEVWorld\cite{zhang2024bevworld} (25.72\% NDS). These accuracy gains confirm the model reliably predicts future states, ensuring a stable foundation for downstream planning.

\subsection{Ablation Studies}
\label{sec:ablation}

We systematically validate the architectural design of TPS-Drive by progressively introducing its core components, as detailed in Table \ref{tab:ablation_components}. 

\paragraph{Effect of Task-Guided Tokenization.} 
Our baseline utilizes a standard reconstruction-guided VQ-VAE\cite{van2017neural} without task-specific supervision. This approach forces the model to encode massive amounts of task-irrelevant background, leading to representation interference, poor spatial reasoning (17.31\% NDS), and suboptimal closed-loop planning (81.1 EPDMS). By introducing the Task-Guided Primary Tokenizer, we explicitly filter out static backgrounds to focus purely on critical dynamic agents. This "representation purification" yields a +12.72\% absolute increase in forecasting accuracy (reaching 30.03\% NDS) and improves the EPDMS to 83.7. This confirms that isolating safety-critical semantics from raw visual data is foundational for effective reasoning in VLA models.

\paragraph{Effect of Residual Refinement Layers.} 
While the purified primary tokenizer establishes a robust semantic foundation, incorporating $L=4$ residual codebooks substantially enhances the model's ability to capture fine-grained geometric and kinematic details. This hierarchical quantization yields an additional +3.75\% absolute boost in spatial forecasting accuracy (reaching 33.78\% NDS). Because the downstream diffusion planner relies on this enriched spatial vocabulary\cite{jiang2025transdiffuser, zheng2024genad}, open-loop planning safety naturally improves, with the ST-P3 Collision Rate (CR) dropping from 0.28\% to 0.26\% and the L2 error decreasing to 0.61m.

\paragraph{Effect of Reward-Driven Refinement.} 
The final training stage is essential for aligning the generalized spatial representation with strict, real-world driving regulations\cite{tang2025plan,shang2025drivedpo}. Applying policy optimization driven by simulated rewards boosts the NAVSIMv2 EPDMS from 84.1 to 86.7. Furthermore, the open-loop collision rate undergoes a drastic reduction to 0.19\%. Interestingly, this policy-level alignment synergistically improves the upstream world modeling quality (NDS increases to 34.60\%), proving that continuous planning rewards effectively encourage the discrete representations to prioritize safety-critical spatial features.


\begin{figure*}[t!]
    \centering
    \includegraphics[width=1.0\linewidth]{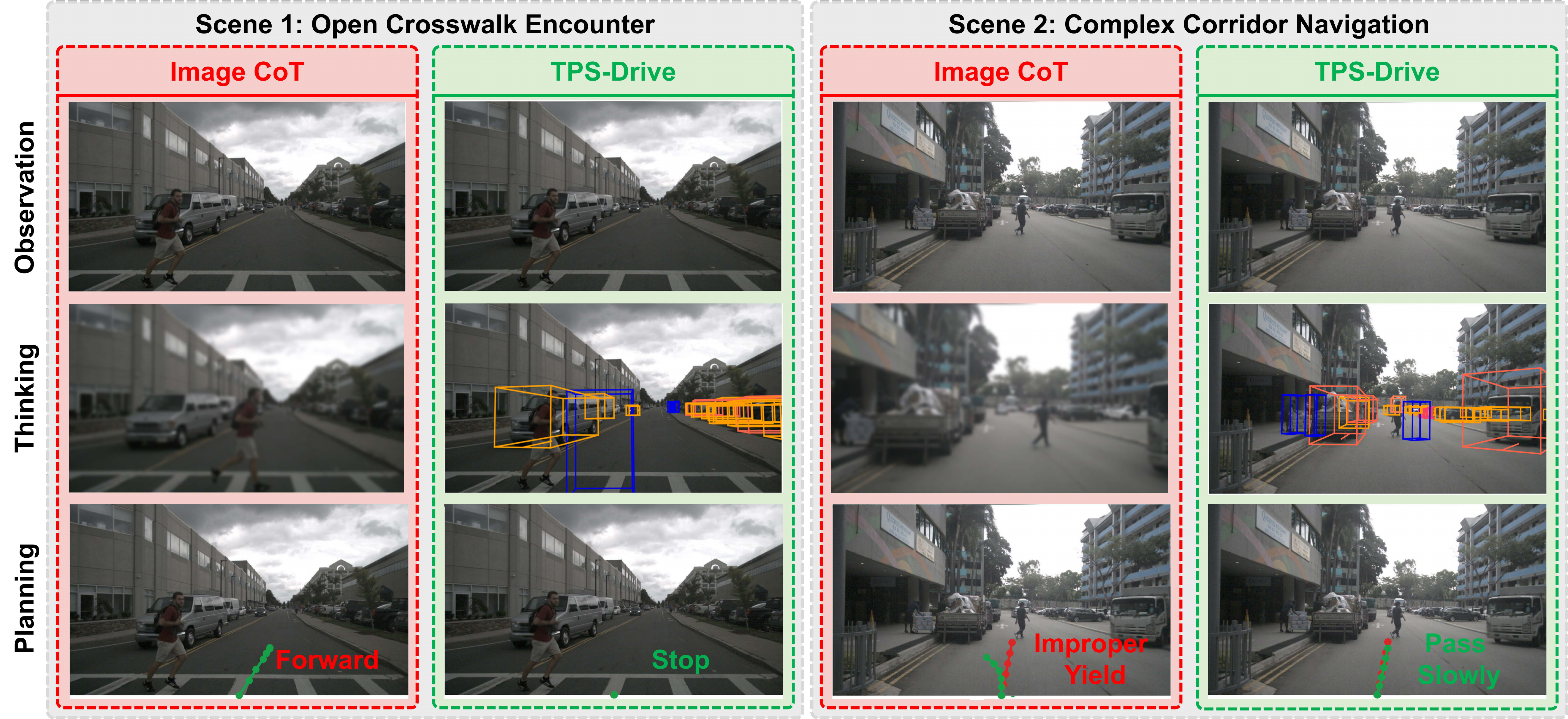} 
    \caption{\textbf{Qualitative comparison in safety-critical scenarios.} Rows denote visual \textbf{Observation}, internal \textbf{Thinking}, and trajectory \textbf{Planning}. In Planning, \textbf{\textcolor{darkgreen}{green}} and \textbf{\textcolor{red}{red}} indicate predicted and ground-truth trajectories, respectively. \textbf{Left:} Facing a pedestrian, the baseline wrongly plans "Forward", while TPS-Drive explicitly predicts 3D bounding boxes to safely "Stop". \textbf{Right:} In a cluttered corridor, the baseline makes an "Improper Yield", whereas TPS-Drive accurately aligns with the ground truth to "Pass Slowly".}
    \label{fig:qualitative_comparison}
\end{figure*}

\subsection{Qualitative Analysis}
Figure \ref{fig:qualitative_comparison} contrasts TPS-Drive with an implicit Image Chain-of-Thought (CoT) baseline \cite{zeng2025futuresightdrive} in safety-critical driving scenarios. Relying on dense image representations, the baseline model suffers from severe representation interference. Overwhelmed by redundant background noise, its semantic reasoning fails to accurately identify task-critical dynamic hazards, leading to dangerous maneuvers. Specifically, the baseline dangerously plans to drive "Forward" into a crossing pedestrian (Scene 1) and executes an "Improper Yield" in a cluttered corridor (Scene 2).

Conversely, TPS-Drive overcomes these limitations by explicitly forecasting the 3D bounding boxes of surrounding agents. By reasoning within this task-guided, purified spatial vocabulary, our planner accurately interprets complex geometric constraints. It successfully isolates critical dynamic agents from background redundancy to synthesize safe and compliant trajectories, such as executing a timely "Stop" (Scene 1) and deciding to "Pass Slowly" (Scene 2). This confirms that explicit tokenization significantly enhances both interpretability and driving safety.

\section{Conclusion}
We propose TPS-Drive, a decoupled framework designed to overcome spatial hallucinations and representation interference in VLM-based autonomous driving. By employing a frozen 3D detection head to supervise the task-guided vector quantization process, our Agent-Centric Tokenizer distills critical dynamic agents from redundant backgrounds into a purified spatial vocabulary. Consequently, TPS-Drive delivers exceptional open-loop spatial forecasting accuracy and establishes new safety records on closed-loop interactive benchmarks, validating that purified geometric representations effectively bridge the gap between high-level semantic reasoning and safe trajectory generation.

\textbf{Limitations and Future Work.} Relying on a frozen detection head bounds representation potential, and the decoupled multi-stage architecture currently restricts real-time inference speed. Future work will explore end-to-end differentiable spatial purification and optimize deployment on dedicated hardware accelerators for low-latency autonomous control.





\bibliographystyle{unsrt}
\bibliography{references}   


\end{document}